\documentclass{vgtc}    
\graphicspath{{figures/}{pictures/}{images/}{./}} 
\usepackage{times}                     
\usepackage{tabu}                      
\usepackage{booktabs}                  
\usepackage{lipsum}                    
\usepackage{mwe}                       
\usepackage{mathptmx}                  

\usepackage{colortbl}
\usepackage{graphicx}                   
\usepackage{subcaption}                 
\usepackage{multirow}                   
\usepackage{amsmath}                    
\onlineid{0}

\vgtccategory{Research}

\vgtcinsertpkg

\title{Accelerated Volumetric Compression without Hierarchies: A Fourier Feature Based Implicit Neural Representation Approach}

\author{Leona Zurkova\\\parbox{1.4in}{\scriptsize \centering leona.zurkova@vsb.cz} %
\and Petr Strakos %
\and Michal Kravcenko %
\and Tomas Brzobohaty%
\and Lubomir Riha %
}
\affiliation{\scriptsize IT4Innovations, VSB – Technical University of Ostrava, Ostrava, Czech Republic}
\teaser{
  \centering
  \includegraphics[width=\linewidth]{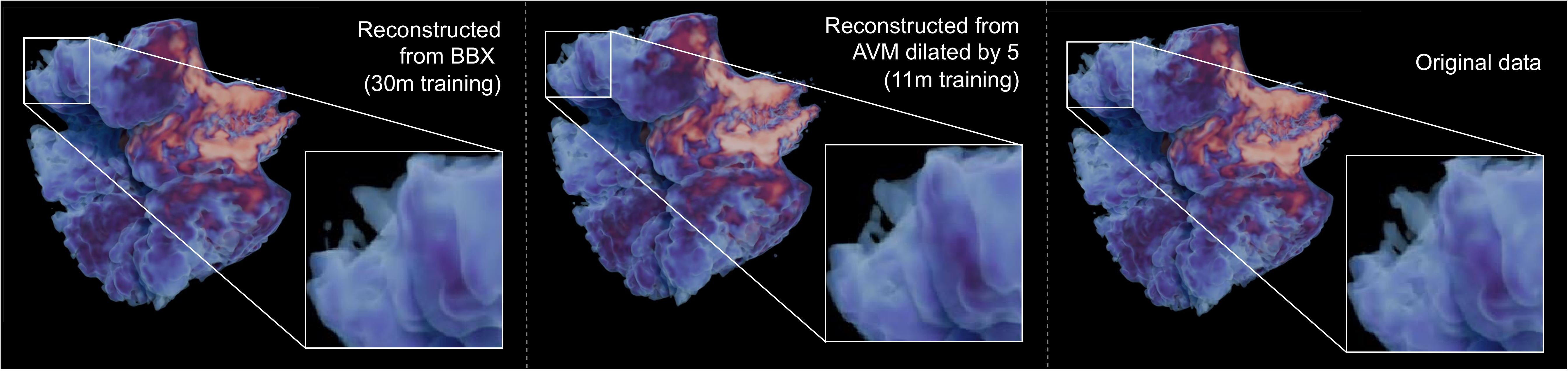}
        \caption{Visual comparison (from left to right) of reconstructed data trained on the entire bounding box, reconstructed data trained on masked active voxel map  dilated by 5 neighbours, and original data.}
        \label{fig:gas_visual} \label{teaser}
}

\abstract{
Volumetric data compression is critical in fields like medical imaging, scientific simulation, and entertainment. We introduce a structure-free neural compression method combining Fourier-feature encoding with selective voxel sampling, yielding compact volumetric representations and faster convergence. Our dynamic voxel selection uses morphological dilation to prioritize active regions, reducing redundant computation without any hierarchical metadata. In the experiment, sparse training reduced training time by 63.7\ \% (from 30 to 11 minutes) with only minor quality loss: PSNR dropped 0.59\,dB (from 32.60 to 32.01) and SSIM by 0.008 (from 0.948 to 0.940). The resulting neural representation, stored solely as network weights, achieves a compression rate of 14  and eliminates traditional data-loading overhead. This connects coordinate-based neural representation with efficient volumetric compression, offering a scalable, structure-free solution for practical applications.

} 


\begin{document}

\maketitle
\section{Introduction}
\label{sec:intro}
Volumetric datasets in fields like Computational Fluid Dynamics (CFD), medical imaging, and entertainment can span hundreds of millions of grid points. Implicit Neural Representations (INRs) compress such data by learning mappings from coordinates $(x, y, z)$ to scalar values, with Fourier features~\cite{tancik2020} improving the modeling of fine detail. NeuralVDB~\cite{kim2024} introduces a hierarchical sparse encoding for high-resolution volumes, and CoordNet~\cite{han2023coordnet} applies coordinate-based networks to various volumetric tasks. Unlike these structure-aware methods, we propose a structure-free approach that accelerates training by sampling only informative voxels. By combining Fourier-feature encoding with dynamic voxel selection from a binary mask, our method enables efficient training on dense tensors while preserving reconstruction quality.

\section{Methodology}
The pipeline starts with normalizing the values per grid and decomposing the source volumetric data into a flattened dataset in several variations: the entire bounding box ($BBX$), active voxel mask ($AVM$), and binary dilated AVM by $\ell$ ($AVM_{\text{dilated}}^{(\ell)}$). The generated datasets are then used to map the right input coordinates into output variables using a Fourier feature-based neural network. 
For testing the neural network inference, the whole bounding box was always used to obtain qualitative metrics from all voxels, even if the network was not trained on them. The results are then compared visually and numerically using metrics described in \ref{sec:metrics}. The pipeline uses Python libraries NumPy and PyTorch, and exploits multi-GPU training on multiple nodes of the LUMI-G supercomputer cluster.

\subsection{Coordinate-based neural network representation}\label{sec:nn_repre}
Our network consists of two distinct parts between the input and output layer. The first part (FF) maps $(x,y,z)$ coordinates to a higher-dimensional Fourier feature space. After this, the FF part passes the mapped coordinates through a multilayer perceptron (MLP). After a brief systematic parameter search (grid search), we estimate the optimal Fourier feature count $k$, hidden MLP layers $m$, $n$ neurons, and other parameters for the selected dataset as stated in~\cref{tab:nn_params}.

\subsection{Fourier feature mapping}\label{sec:fourier}
Following~\cite{tancik2020}, we prepared sinusoidal mapping
\begin{equation}
FF(\mathbf{p}) = [\cos(2\pi\mathbf{B}\mathbf{p}), \sin(2\pi\mathbf{B}\mathbf{p})],
\end{equation}
where each input coordinate vector $\mathbf{p}=(x,y,z)$ is mapped to a set of sine and cosine functions with different frequencies, allowing the network to learn both low and high-frequency components more effectively. We construct $\mathbf{B}=(\mathbf{B}_1, \mathbf{B}_2, ..., \mathbf{B}_n)^T$, where each $B_i$ is sampled from zero‐mean Gaussian distributions, each scaled by a distinct learnable \textit{gauss multiplier} to cover multiple frequency bands.

\subsection{Selective training strategy}
We used different variations of training data to train the neural network. One variation, which we labeled as $BBX$, contains all voxels from the entire bounding box, including zero-valued ones in the background. Another one is a subset of $BBX$, an active voxel mask ($AVM$) defined as 
\begin{equation}
    AVM(x,y,z) = 
    \begin{cases} 
        1, & \text{if } \exists n : D(x,y,z,n) > 0 \\ 
        0, & \text{otherwise},
    \end{cases}
    \label{eq:mask}
    \end{equation}
which is active per voxel if at least one grid variable value $n$ on coordinates $(x, y, z)$ of the normalized data $D$ has a positive value. 

The last variation to mention is $\ell\times$ dilated $AVM$, which in addition includes $\ell$ zero-valued voxels around the original $AVM$ and is defined recursively as
\begin{equation}
AVM_{\text{dilated}}^{(\ell)}(x,y,z) = \bigcup_{(r, s, t) \in \{-1,0,1\}^3} AVM_{\text{dilated}}^{(\ell-1)}(x+r,y+s,z+t).
\label{eq:dilation}
\end{equation}
 Here we evaluate only cases $\ell\in\{1,2,5,10\}$, and special cases can be considered as $AVM_{\text{dilated}}^{(0)}=AVM$, and $AVM_{\text{dilated}}^{(\infty)}=BBX$. Dilated masks provide a controllable 'halo' of context around active regions, trading off training size against boundary smoothness.

\subsection{Dataset and preprocessing}

We demonstrate our selective voxel masking strategy on results from CFD simulation in VDB format with dimensions 239$\times$239$\times$387. The size of all values in single precision is 125 MB, and the original VDB file is 43 MB. The file resembles a gas burner fire consisting of a density and temperature grid. The density grid can be seen rendered in Figure \ref{fig:gas_visual}.

\subsection{Metrics}
\label{sec:metrics}
We assess reconstruction quality using three standard image similarity metrics: normalized root mean squared error (NRMSE), peak signal-to-noise ratio (PSNR), and structural similarity index measure (SSIM). These capture both pixel-level accuracy and perceptual similarity. 

To make the model comparison more intuitive, we define a cumulative quality metric
\begin{equation}
\scalebox{1}{$
    \text{Score} =  \frac{1}{3} \left(\frac{\text{PSNR}}{MAX_{PSNR}} + (1 - \text{NRMSE}) + \text{SSIM}\right) \in \langle0, 1\rangle, \,
     $}
    \label{eq:score}
\end{equation}
which equally combines the normalized contributions of PSNR, NRMSE, and SSIM. The SSIM and NRMSE values are already in the range of [0,1], however, the PSNR is given in dB, so a maximum value $\text{PSNR}_{MAX} = 48 dB$ was chosen for normalization. $\text{Score}$ ranges from 0 (poor) to 1 (perfect) and summarizes overall quality. \textbf{Theoretical compression} is calculated as the dataset value count divided by the number of neural network weights.

\section{Results}\label{sec:results}
The set of hyperparameters utilized in this experiment is documented in Table \ref{tab:nn_params}.

\begin{figure}[ht]
    \centering
    \begin{subfigure}[b]{0.29\columnwidth}
        \centering
        \includegraphics[width=\columnwidth]{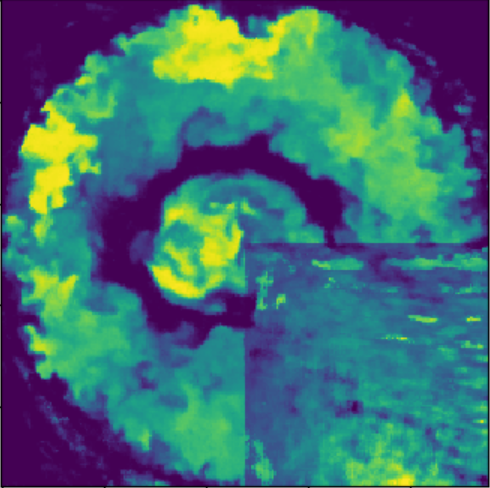}
        \caption{$AVM$}
        \label{fig:sub-a}
    \end{subfigure}
    \begin{subfigure}[b]{0.29\columnwidth}
        \centering
        \includegraphics[width=\columnwidth]{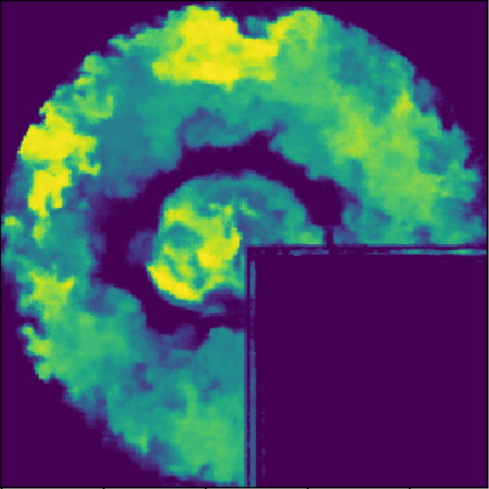}
        \caption{$AVM_{\text{dilated}}^{(1)}$}
        \label{fig:sub-b}
    \end{subfigure}
    \begin{subfigure}[b]{0.29\columnwidth}
        \centering
        \includegraphics[width=\columnwidth]{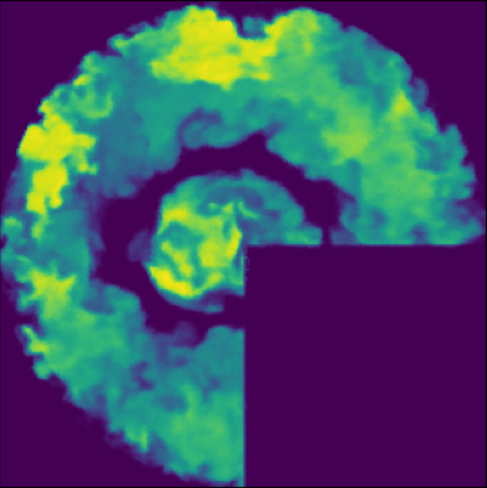}
        \caption{$BBX$}
        \label{fig:sub-f}
    \end{subfigure}

    \caption{Visualizations of 200th slice of Z axis of the density grid from the result for networks trained on (a) $AVM$, (b) $AVM$ dilated by 1, and (c) $BBX$ data.}
    \label{fig:grid-figure}
\end{figure}

\begin{table}[]
    \caption{Hyperparameter values used for this experiment: epochs~$e$, number of fourier features~$k$, gauss multiplier~$gm$, number of hidden layers~$m$, number of neurons in those layers~$n$, batch size~$bs$, and learning rate~$lr$.}
    \centering
    \begin{tabular}{rrrrrrr|r}
     \multicolumn{1}{c}{\textbf{$e$}} & \multicolumn{1}{c}{\textbf{$k$}} &\multicolumn{1}{c}{\textbf{$gm$}} & \multicolumn{1}{c}{\textbf{$m$}} & \multicolumn{1}{c}{\textbf{$n$}} & \multicolumn{1}{c}{\textbf{$bs$}} & \multicolumn{1}{c|}{\textbf{$lr$}} & \multicolumn{1}{c}{compression} \\ \hline
     100 & 1280 &2.5& 8 & 512 & 24 & 0.0003 & 14.0
    \end{tabular}\label{tab:nn_params}
    \end{table}
 The numeric comparison for each dilation variation can be seen in Figure~\ref{tab:results_metrics}. A visual comparison between the original data, the densely trained ($BBX$) reconstruction and the sparsely trained reconstruction ($AVM$ dilated by 5) can be seen in Figure \ref{fig:gas_visual}. Visual artifacts for $AVM$ and its minimally dilated version can be seen in Figure \ref{fig:grid-figure}.
 
\begin{table}[]
    \caption{Results for parameters in Table \ref{tab:nn_params}. Gray lines are the rendered results in Figure \ref{fig:gas_visual}.}

    \begin{tabular}{l|rrrrr}
    \multicolumn{1}{c|}{} & \multicolumn{1}{c}{\textbf{PSNR (dB)}} & \multicolumn{1}{c}{\textbf{NRMSE}} & \multicolumn{1}{c}{\textbf{SSIM}} & \multicolumn{1}{c}{\textbf{Score}} & \multicolumn{1}{c}{\textbf{Time (s)}} \\ \hline
    AVM & 15.47 & 0.748 & 0.536 & 0.37 & 554 \\
    $\ell=1$ & 29.11 & 0.120 & 0.917 & 0.80 & 585 \\
    $\ell=2$ & 28.41 & 0.128 & 0.916 & 0.79 & 576 \\
    \rowcolor[HTML]{C0C0C0} 
    $\ell=5$ & 32.01 & 0.085 & 0.940 & 0.84 & 646 \\
    $\ell=10$ & 32.21 & 0.084 & 0.947 & 0.84 & 693 \\
    \rowcolor[HTML]{C0C0C0} 
    BBX & 32.60 & 0.081 & 0.948 & 0.85 & 1780
    \end{tabular}
    \label{tab:results_metrics}
    \end{table}

\section{Discussion and Conclusion}
As Table~\ref{tab:results_metrics} shows, reconstruction quality is comparable when training on the full bounding box ($BBX$) and on dilated active voxel masks ($AVM_{dilated}$), but $BBX$ takes much longer. However, using only the basic active-voxel map (AVM) can fail if the dataset lacks smooth transitions between active and background regions. For example, in our gas-burner dataset with a missing quadrant, the network hallucinates the absent region (see Figure~\ref{fig:sub-a}). Adding a minimal dilation like $\ell=1$ then produces artifacts at sharp edges. More aggressive dilation may help, though optimal size remains dataset-dependent.

Future work includes pruning strategies for faster training, exploring non-MLP architectures, and refining hyperparameter tuning to balance compression and reconstruction quality. These improvements could broaden applicability to diverse volumetric data and visualization scenarios.

\acknowledgments{
This work was supported by the Ministry of Education, Youth and Sports of the Czech Republic through the e-INFRA CZ (ID:90254). This article has been produced with the financial support of the European Union under the REFRESH – Research Excellence For REgion Sustainability and High-tech Industries project number CZ.10.03.01/00/22\_003/0000048 via the Operational Programme Just Transition.
}

\bibliographystyle{abbrv-doi}

\bibliography{template}
\end{document}